\documentclass[runningheads]{llncs}
\usepackage{graphicx}
\usepackage{amsmath}

\usepackage{textcomp}
\usepackage{xcolor}
\usepackage{algorithm}
\usepackage{amsmath}
\usepackage{amssymb}
\usepackage{algpseudocode} 
\usepackage{caption}
\usepackage{subcaption}

\usepackage{multirow}
\usepackage{tabularx}
\usepackage{longtable}
\usepackage{booktabs}
\begin{document}
\title{Fusion LiDAR-Inertial-Encoder data for High-Accuracy SLAM}
%
%
\author{Manh Do Duc, Thanh Nguyen Canh \and Minh DoNgoc \and Xiem HoangVan$^*$}
\authorrunning{Xiem H. et al.}
%
\institute{University of Engineering and Technology, Vietnam National University \\
Hanoi, Vietnam (ddmanh99@gmail.com, \{canhthanh, dongocminh, xiemhoang\}@vnu.edu.vn) \\
$^*$ Corresponding author
}

\maketitle              
\begin{abstract}
In the realm of robotics, achieving simultaneous localization and mapping (SLAM) is paramount for autonomous navigation, especially in challenging environments like texture-less structures. This paper proposed a factor-graph-based model that tightly integrates IMU and encoder sensors to enhance positioning in such environments. The system operates by meticulously evaluating the data from each sensor. Based on these evaluations, weights are dynamically adjusted to prioritize the more reliable source of information at any given moment. The robot's state is initialized using IMU data, while the encoder aids motion estimation in long corridors. Discrepancies between the two states are used to correct the IMU drift. The effectiveness of this method is demonstrably validated through experimentation. Compared to Karto SLAM, a widely used SLAM algorithm, this approach achieves an improvement of 26.98\% in rotation angle error and a reduction of 67. 68\% in position error. These results convincingly demonstrate the method's superior accuracy and robustness in textureless environments.

\end{abstract}
\section{Introduction}

The field of Autonomous Mobile Robots (AMRs) is experiencing a surge of development across diverse industries such as healthcare, security monitoring~\cite{phala2016air},~\cite{hancke2010security}, and construction~\cite{nubert2022graph}. To thrive in these dynamic environments and fulfill their designated tasks, AMRs necessitate robust capabilities in three critical areas: information gathering~\cite{matia1998multisensor}, environment perception~\cite{guo2006environmental}, and precise Simultaneous Localization and Mapping (SLAM)~\cite{li2012overview},~\cite{canh2024s3m},~\cite{canh2023object}. Although various research efforts have explored effective mapping techniques that utilize individual sensors such as LiDAR~\cite{tee2021lidar} or cameras~\cite{servieres2021visual}, cameras have significant limitations. Their sensitivity to light fluctuations can considerably hinder reliable operation. To address this challenge, this paper delves into a LiDAR-based approach for SLAM.  LiDAR offers substantial advantages, including a marked reduction in susceptibility to variations in lighting conditions and the ability to operate at high frequencies. However, established 2D LiDAR-based mapping methods, such as GMapping and Hector SLAM, showcase success primarily in controlled indoor environments~\cite{zhao2022research}. These methods struggle to adapt to environments characterized by sparsity, featuring long corridors, tunnels, or open spaces. 

To address these limitations of single sensors, recent research has explored multi-sensor fusion techniques by leveraging the strengths of complementary sensors. LIO-SAM~\cite{shan2020lio} integrates an IMU not only to predict robot motion prior to LiDAR scan matching, but also to initiate the LiDAR odometry optimization process. This combined approach refines IMU sensor error estimates within a factor graph. Several previous works~\cite{wang2021gr},~\cite{canh2022multisensor} proposed a multimodal system that integrates LiDAR, camera, IMU, encoder, and GPS to estimate the robust and accurate state of the robot. It can also eliminate dynamic targets and unstable features. Overall, tightly coupled multi-sensor fusion systems are becoming essential for ensuring the accuracy and flexibility required for autonomous robots operating in complex environments.

This paper proposes an effective method that tightly integrates data from multiple sensors, including LiDAR, IMUs, and encoders, to estimate robot states using factor graphs. Our key advantages include: (1) sensor uncertainty detection, which is estimated when the robot moves in a new scenario with each sensor, (2) adaptive noise assignment- it can adjust for noise levels based on the sensor uncertainty detection part, and (3) reconfigurable optimization, which can automatically reconfigure the pose graph optimization process throughout robot operation. We evaluated our proposed method through physical simulations, demonstrating its accuracy and applicability in the real environment. 
 
\section{Methodology} \label{sec:method}

\begin{figure}
    \centering
    \includegraphics[width=1\linewidth]{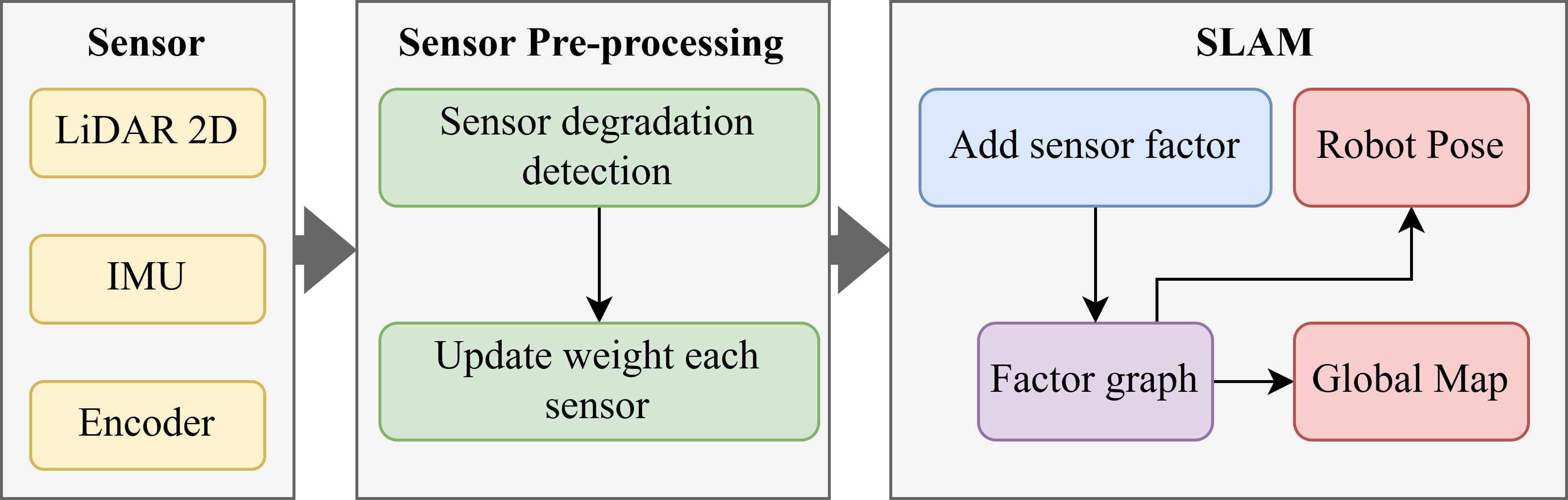}
    \caption{Pipeline of our method.}
    \label{fig:overview}
\end{figure}

The pipeline of the method is shown in Fig.~\ref{fig:overview} and contains two nodes. Our robot utilizes three sensors: 2D LiDAR generates a 2D point cloud, providing a detailed representation of the surrounding environment; the IMU measures the robot’s angular velocity and linear acceleration, aiding in motion estimation; the encoder tracks the distance traveled by the robot. Sensor preprocessing involves data caching and synchronization to ensure consistency. Raw IMU data are used to estimate robot motion and pre-integrated according to the time stamp to remove the motion distortion of the point cloud. In addition, the sensor can detect degradation and update the weight during movement. Robot state estimation employs a maximum a posteriori (MAP) approach with a factor graph to model robot position and trajectory based on sensor measurements. Finally, a 2D map of the environment is constructed using the processed LiDAR data. 

Although multisensor systems offer improved positioning accuracy by incorporating various sensors, it is crucial to strike a balance. An increase in the number of sensors directly results in higher computational complexity. In addition, not all sensors perform well in every environment. Including an inappropriate sensor can negatively impact the state estimation process. Therefore, careful sensor selection is vital for multi-sensor systems.
Our experiments revealed that LiDAR performance degrades in highly repetitive environments such as corridors and tunnels. To address this, we analyze the distance difference between consecutive LiDAR scans. If this difference falls below a predefined threshold, it indicates potential LiDAR degradation. Wheel slippage, a common occurrence during movement, leads to accumulated errors in encoder data. We mitigate this by comparing the IMU-predicted position change with the encoder data. Discrepancies exceeding a set threshold exclude the encoder data from the system. IMU data, while relatively accurate in the short term, suffer from accumulating bias over time. To address this, we compare two consecutive estimated states to correct for the IMU bias.

\begin{figure}
    \centering
    \includegraphics[width=1\linewidth]{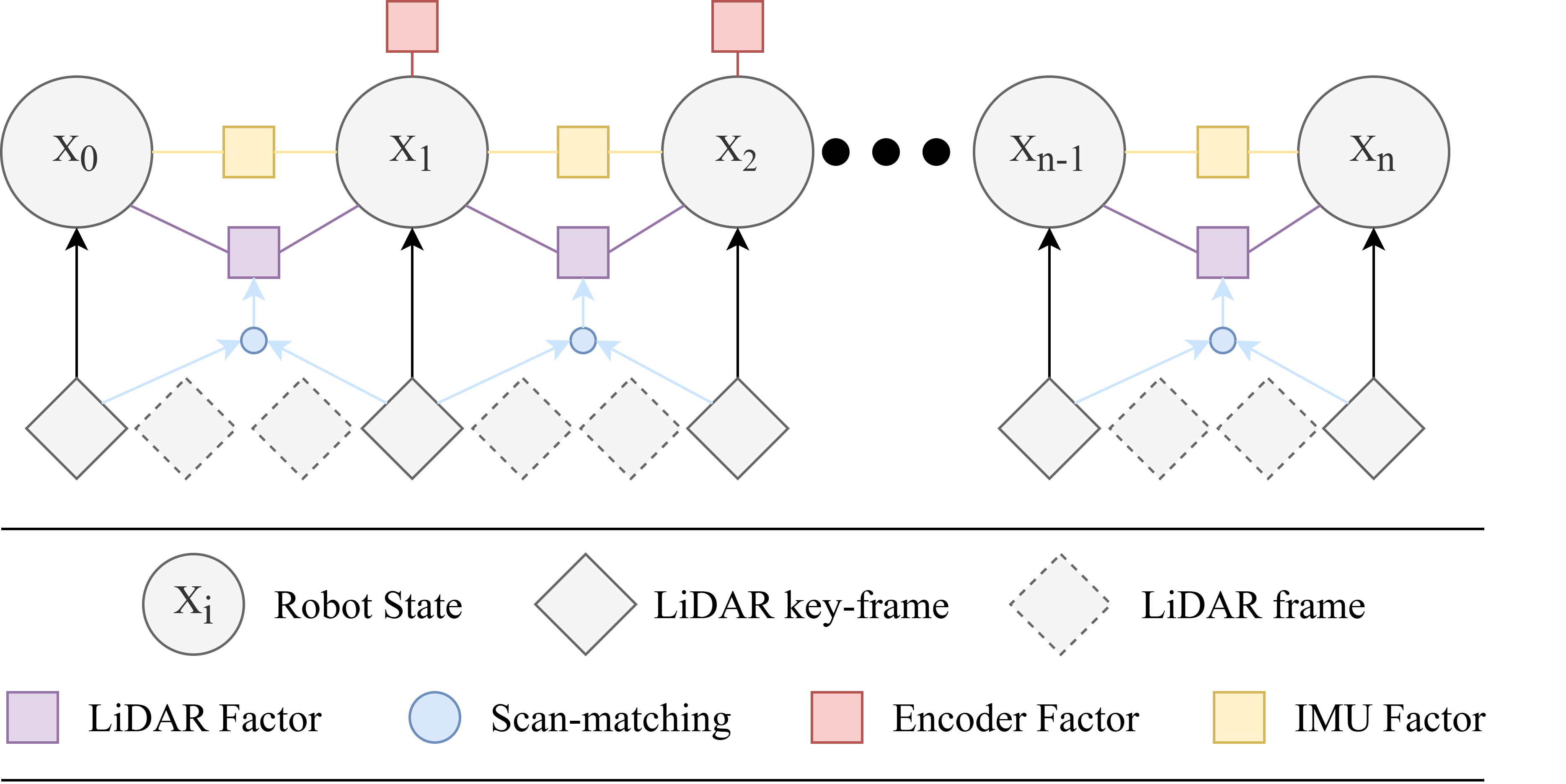}
    \caption{Our sensor system is based on a graph.}
    \label{fig:fg}
\end{figure}

Our approach leverages a factor graph to represent the relationships between sensor measurements and robot states (Fig.~\ref{fig:fg}). Each robot state corresponds to a node in the graph, connected to other states by constraints. These constraints come in the following forms: LiDAR odometry factor, IMU factor, and encoder factor. If a sensor exhibits performance degradation, the corresponding constraint is excluded from the factor graph to maintain accuracy. Adding constraints to the system is like adding noise to the sensor measurements. The weight depends on how far the measurements deviate from a threshold. In case of high repeatability detection from laser data, the constraint from LiDAR odometry will not be added to the graph. 

Assuming the coordinate frames of the LiDAR, IMU, and encoder are aligned with the robot’s coordinate frame. The robot’s state is expressed as a vector as shown in equation~(\ref{eq:x}). In this equation, $r$ represents the rotation matrix, $t$ represents the translation matrix, $v$ represents the velocity matrix, and b represents the IMU bias.

\begin{equation} \label{eq:x}
    x = \begin{bmatrix}
        r^T & t^T & v^T & b^T
    \end{bmatrix}
\end{equation}

The raw data of IMU includes the angular velocity and linear acceleration as shown in equation~(\ref{eq:placeholder}, \ref{eq:placeholder_label}). These measurements are affected by bias $b$ and noise $n$.

\begin{equation}
    \omega_t = \overline{\omega}_t + b_t^{\omega} + n_t^{\omega}
    \label{eq:placeholder}
\end{equation}

\begin{equation}
    a_t = r_t \overline{a}_t + b_t^a + n_t^a
    \label{eq:placeholder_label}
\end{equation}

The robot’s state, including its velocity, position, and orientation, is then estimated by IMU data as follows:
\begin{equation}
    v_{t+\delta}^I = v_t + r_t^T a_t \delta
    \label{eq:v}
\end{equation}

\begin{equation}
    p_{t+\delta}^i = p_t + v_t \delta + \frac{1}{2} r_t^T a t \beta^2 
    \label{eq:p}
\end{equation}

\begin{equation}
    r_{t+\delta}^l = r_t^T e^{\omega_t ^\delta}
    \label{eq:r}
\end{equation}

The IMU provides constraint position information between two consecutive states, which is calculated as follows:
\begin{equation}
  \Delta_v^i = r_t^T \left( v_{t+\delta} - v_t \right)
  \label{eq:dv}
\end{equation}

\begin{equation}
\Delta_p^i = r_t^T \left( p_{t+\delta} - p_t \right)
\label{eq:dp}
\end{equation}

\begin{equation}
   \Delta_r^i = r_t^T r_{t + \delta}
   \label{eq:dr}
\end{equation}

The encoder provides the displacement between two states. The scan matching step of LiDAR also estimates the displacement of the two states. The position constraints of the encoder and LiDAR are as follows:

\begin{equation}
    \Delta_p^0 = r_t^T \left( p_{t+\delta} - p_t \right)
    \label{eq:d0p}
\end{equation}

\begin{equation}
    \Delta_p^L = r_t^T \left( p_{t+\delta} - p_t \right)
    \label{eq:elp}
\end{equation}
These constraints are added to the factor graph along with the weights selected from the sensor degradation evaluation step above. Whenever a new state and its corresponding constraints are added to the graph, the iSAM2~\cite{kaess2012isam2} optimization method is employed to calculate the most likely robot state based on all available constraints.

\section{Experimental Results}

\begin{figure}[!ht]
    \centering
    \begin{subfigure}[b]{0.45\textwidth}
    \centering
    \includegraphics[width=\textwidth]{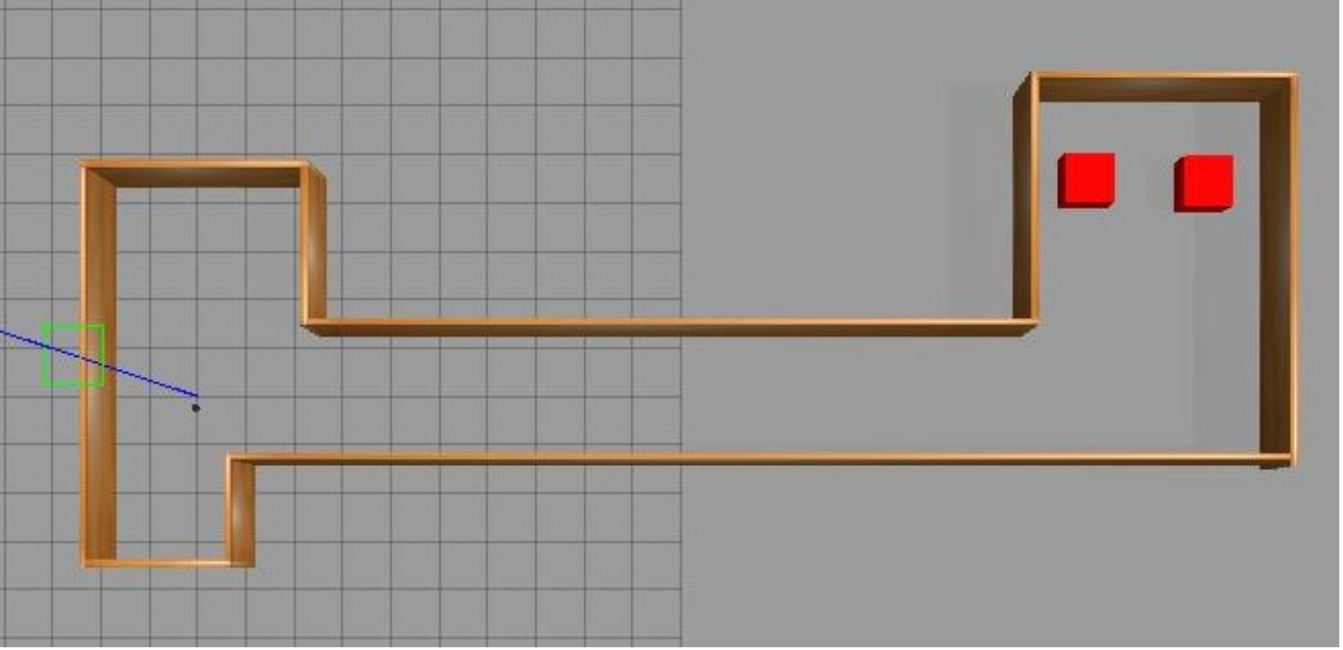}
    \caption{Environment in Gazebo}
    \label{fig:env}
    \end{subfigure}
    \centering
    \begin{subfigure}[b]{0.45\textwidth}
    \centering
    \includegraphics[width=\textwidth]{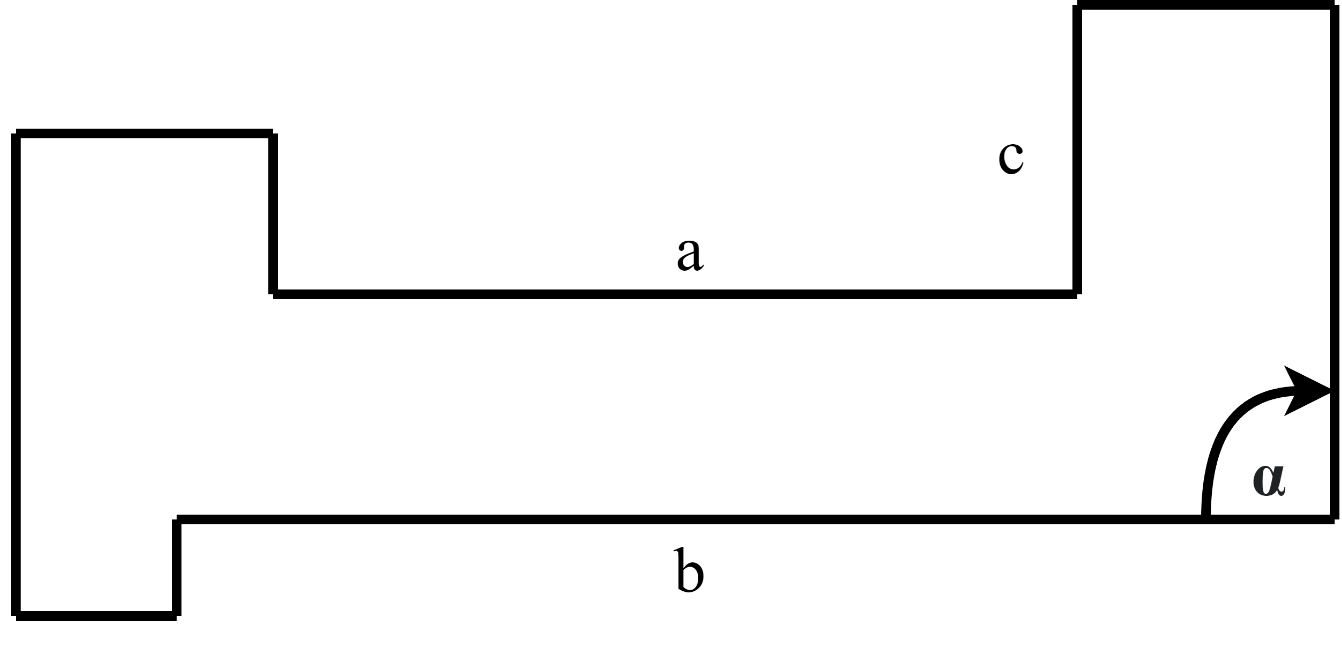}
    \caption{Environment with key distances.}
    \label{fig:map}
    \end{subfigure}
    \caption{Simulation Environment.}
    \label{fig:si}
\end{figure}

The simulated corridor environment was created in Gazebo (Fig.~\ref{fig:env}). This environment presented a challenge for 2D LiDAR-based SLAM methods due to the repetitive nature of corridors, which can lead to drift problems during the scan-matching process. We defined key distance variables on the map as shown in Fig.~\ref{fig:map} to quantify the accuracy of the map.

\begin{figure}
    \centering
    \includegraphics[width=1\linewidth]{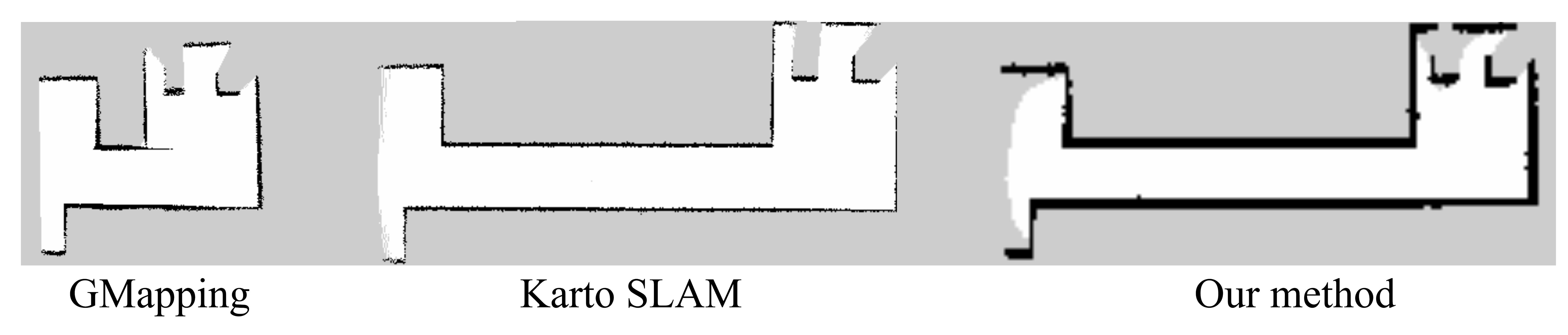}
    \caption{Mapping result by (a) GMapping, (b) Karto SLAM, (c) Our method.}
    \label{fig:results}
\end{figure}

\begin{figure}
    \centering
    \includegraphics[width=0.9\linewidth]{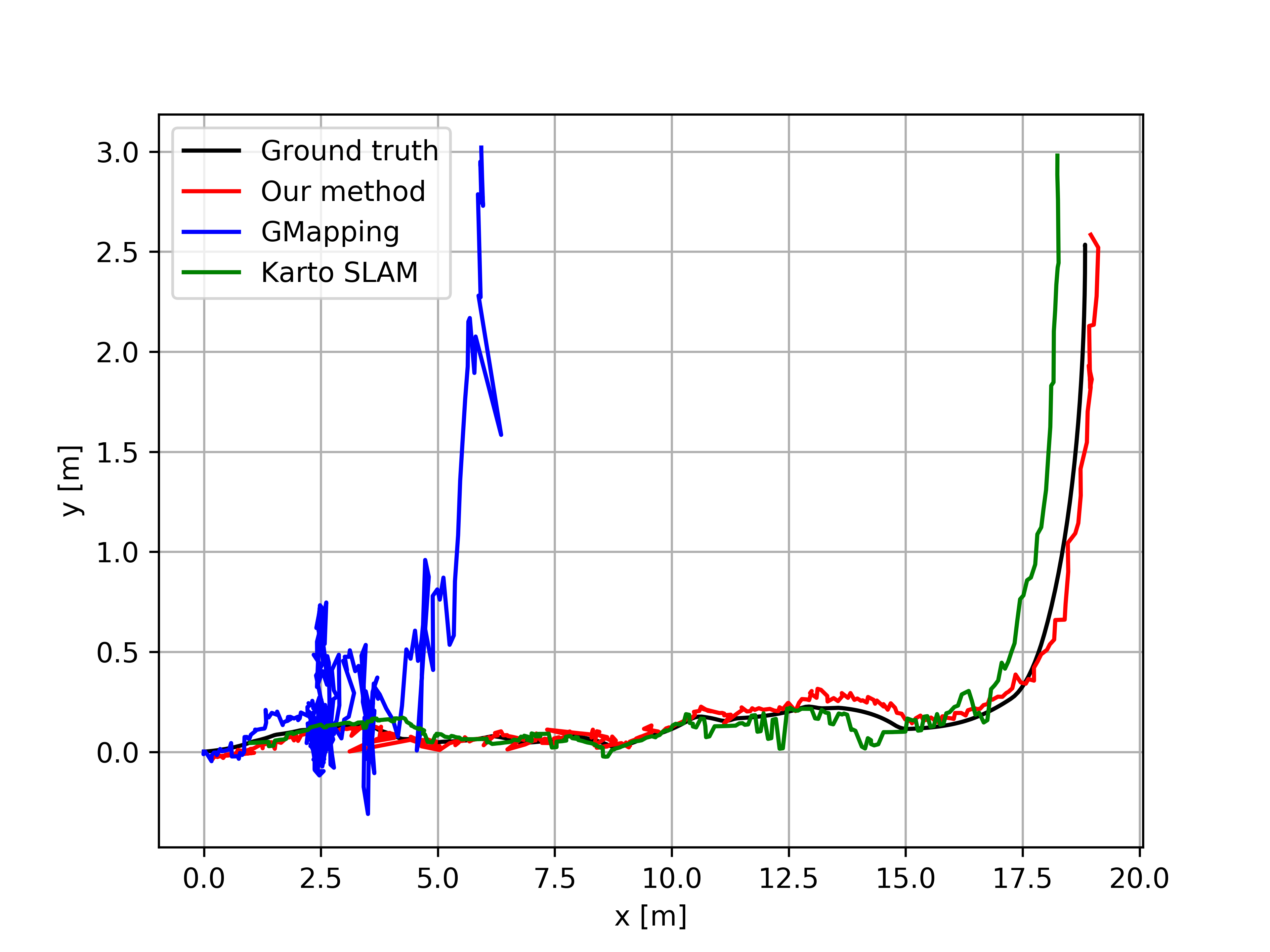}
    \caption{Trajectory comparison.}
    \label{fig:traj}
\end{figure}

To evaluate the performance of different SLAM algorithms, we guided the robot through a corridor while acquiring sensor data, including LiDAR, IMU, and encoder readings. Subsequently, we implemented three SLAM algorithms and assessed their accuracy in generating a map of the environment and tracking the robot's trajectory. The accuracy of the map is presented in Table~\ref{tab:comparison}, which demonstrates that our method exhibits reduced robot drift compared to Karto SLAM when moving in the corridor environment, mapping results by three methods presented in Fig.~\ref{fig:results}, and it can be seen that the map from the GMapping algorithm is very inaccurate. In addition, we observe a partial improvement in the map deviation. Fig.~\ref{fig:traj} shows the comparison of the robot’s trajectory during movement. We utilized the RMSE metric to quantify the accuracy of both the position and the rotation errors. 

\begin{table}[h]
\centering
\caption{Comparison of Variables and Methods}
\begin{tabular}{p{0.15\textwidth}|p{0.25\textwidth}|p{0.25\textwidth}|p{0.25\textwidth}}
\hline
\textbf{Variable} & \textbf{True Value} & \textbf{Karto SLAM} & \textbf{Our method} \\ \hline
a                 & 14.500              & 13.415              & \textbf{13.897}     \\
b                 & 21.000              & 19.600              & \textbf{20.226}     \\
c                 & 5.000               & 4.863               & \textbf{4.908}      \\
$\alpha$          & 90.00               & 92.286              & \textbf{89.236}     \\ \hline
\end{tabular}
\label{tab:comparison}
\end{table}

Our evaluation compared two established SLAM methods: Gmapping~\cite{grisetti2005improving} and KartoSLAM~\cite{konolige2010efficient}. Gmapping utilizes a particle filter, where each particle represents a robot state and map. However, in highly repetitive environments, particles may not update significantly, leading to inaccurate robot pose estimation and substantial errors. Karto SLAM is a graph-based SLAM technique that initializes the robot state using odometry data. While subsequent LiDAR matching refines the pose the reliance on odometry, which is prone to slip and large rotation angle error can lead to significant errors in the robot’s directional estimate. Based on graph optimization, our approach uses additional data from the IMU sensor to improve the estimation of the rotation angle of the robot by 26\% compared to the Karto SLAM as shown in Table~\ref{tab:slam-comparison}. Furthermore, we incorporate odometry data and IMU data to provide additional constraints, resulting in a more accurate overall estimate of robot state.

\begin{table}[h]
    \centering
    \caption{Comparison of Position and Angle Errors of Different SLAM Methods}
    \begin{tabular}{p{0.25\textwidth}|p{0.35\textwidth}|p{0.35\textwidth}}
        \hline
        \textbf{Method} & \textbf{Error of position (m)} & \textbf{Error of angle (degree)} \\
        \hline
        \textbf{GMapping} & 4.5489 & 1.357 \\
        \hline
        \textbf{Karto SLAM} & 0.6032 & 0.923 \\
        \hline
        \textbf{Our Method} & \textbf{0.1949} & \textbf{0.674} \\
        \hline
    \end{tabular}
    \label{tab:slam-comparison}
\end{table}

\section{Conclusion}
This paper presented an effective multisensor fusion approach for SLAM in textureless environments. Our method integrates a sensor selection and weighting strategy that leverages sensor characteristics to prioritize high-quality measurements. This combined approach significantly improves the accuracy of robot positioning and mapping. The factor graph framework enables flexible adaptation and integration of various sensor types within the system. Our experimental evaluation demonstrated the ability of our method compared to Karto SLAM, achieving substantial reductions in both the positioning error (67. 68\%) and the angular error (26. 98\%). Future work will focus on developing a loop closure mechanism, which is crucial for global consistency in SLAM. In addition, we plan to conduct real-world experiments to validate the effectiveness of our approach in practical scenarios.

%
%
\bibliographystyle{IEEEtran}
\bibliography{ref}

\begin{thebibliography}{10}
\providecommand{\url}[1]{#1}
\csname url@samestyle\endcsname
\providecommand{\newblock}{\relax}
\providecommand{\bibinfo}[2]{#2}
\providecommand{\BIBentrySTDinterwordspacing}{\spaceskip=0pt\relax}
\providecommand{\BIBentryALTinterwordstretchfactor}{4}
\providecommand{\BIBentryALTinterwordspacing}{\spaceskip=\fontdimen2\font plus
\BIBentryALTinterwordstretchfactor\fontdimen3\font minus
  \fontdimen4\font\relax}
\providecommand{\BIBforeignlanguage}[2]{{%
\expandafter\ifx\csname l@#1\endcsname\relax
\typeout{** WARNING: IEEEtran.bst: No hyphenation pattern has been}%
\typeout{** loaded for the language `#1'. Using the pattern for}%
\typeout{** the default language instead.}%
\else
\language=\csname l@#1\endcsname
\fi
#2}}
\providecommand{\BIBdecl}{\relax}
\BIBdecl

\bibitem{phala2016air}
K.~S.~E. Phala, A.~Kumar, and G.~P. Hancke, ``Air quality monitoring system
  based on iso/iec/ieee 21451 standards,'' \emph{IEEE Sensors Journal},
  vol.~16, no.~12, pp. 5037--5045, 2016.

\bibitem{hancke2010security}
G.~P. Hancke, K.~Markantonakis, and K.~E. Mayes, ``Security challenges for
  user-oriented rfid applications within the" internet of things",''
  \emph{Journal of Internet Technology}, vol.~11, no.~3, pp. 307--313, 2010.

\bibitem{nubert2022graph}
J.~Nubert, S.~Khattak, and M.~Hutter, ``Graph-based multi-sensor fusion for
  consistent localization of autonomous construction robots,'' in \emph{2022
  International Conference on Robotics and Automation (ICRA)}.\hskip 1em plus
  0.5em minus 0.4em\relax IEEE, 2022, pp. 10\,048--10\,054.

\bibitem{matia1998multisensor}
F.~Mat{\'\i}a and A.~Jim{\'e}nez, ``Multisensor fusion: an autonomous mobile
  robot,'' \emph{Journal of Intelligent and robotic systems}, vol.~22, pp.
  129--141, 1998.

\bibitem{guo2006environmental}
L.~Guo, M.~Zhang, Y.~Wang, and G.~Liu, ``Environmental perception of mobile
  robot,'' in \emph{2006 IEEE International Conference on Information
  Acquisition}.\hskip 1em plus 0.5em minus 0.4em\relax IEEE, 2006, pp.
  348--352.

\bibitem{li2012overview}
J.~Li, L.~Cheng, H.~Wu, L.~Xiong, and D.~Wang, ``An overview of the
  simultaneous localization and mapping on mobile robot,'' in \emph{2012
  Proceedings of International Conference on Modelling, Identification and
  Control}.\hskip 1em plus 0.5em minus 0.4em\relax IEEE, 2012, pp. 358--364.

\bibitem{canh2024s3m}
T.~N. Canh, V.-T. Nguyen, X.~HoangVan, A.~Elibol, and N.~Y. Chong, ``S3m:
  Semantic segmentation sparse mapping for uavs with rgb-d camera,'' in
  \emph{2024 IEEE/SICE International Symposium on System Integration
  (SII)}.\hskip 1em plus 0.5em minus 0.4em\relax IEEE, 2024, pp. 899--905.

\bibitem{canh2023object}
T.~N. Canh, A.~Elibol, N.~Y. Chong, and X.~HoangVan, ``Object-oriented semantic
  mapping for reliable uavs navigation,'' in \emph{2023 12th International
  Conference on Control, Automation and Information Sciences (ICCAIS)}.\hskip
  1em plus 0.5em minus 0.4em\relax IEEE, 2023, pp. 139--144.

\bibitem{tee2021lidar}
Y.~K. Tee and Y.~C. Han, ``Lidar-based 2d slam for mobile robot in an indoor
  environment: A review,'' in \emph{2021 International Conference on Green
  Energy, Computing and Sustainable Technology (GECOST)}.\hskip 1em plus 0.5em
  minus 0.4em\relax IEEE, 2021, pp. 1--7.

\bibitem{servieres2021visual}
M.~Servi{\`e}res, V.~Renaudin, A.~Dupuis, and N.~Antigny, ``Visual and
  visual-inertial slam: State of the art, classification, and experimental
  benchmarking,'' \emph{Journal of Sensors}, vol. 2021, no.~1, p. 2054828,
  2021.

\bibitem{zhao2022research}
J.~Zhao, S.~Liu, and J.~Li, ``Research and implementation of autonomous
  navigation for mobile robots based on slam algorithm under ros,''
  \emph{Sensors}, vol.~22, no.~11, p. 4172, 2022.

\bibitem{shan2020lio}
T.~Shan, B.~Englot, D.~Meyers, W.~Wang, C.~Ratti, and D.~Rus, ``Lio-sam:
  Tightly-coupled lidar inertial odometry via smoothing and mapping,'' in
  \emph{2020 IEEE/RSJ international conference on intelligent robots and
  systems (IROS)}.\hskip 1em plus 0.5em minus 0.4em\relax IEEE, 2020, pp.
  5135--5142.

\bibitem{wang2021gr}
T.~Wang, Y.~Su, S.~Shao, C.~Yao, and Z.~Wang, ``Gr-fusion: Multi-sensor fusion
  slam for ground robots with high robustness and low drift,'' in \emph{2021
  IEEE/RSJ International Conference on Intelligent Robots and Systems
  (IROS)}.\hskip 1em plus 0.5em minus 0.4em\relax IEEE, 2021, pp. 5440--5447.

\bibitem{canh2022multisensor}
T.~N. Canh, T.~S. Nguyen, C.~H. Quach, X.~HoangVan, and M.~D. Phung,
  ``Multisensor data fusion for reliable obstacle avoidance,'' in \emph{2022
  11th International Conference on Control, Automation and Information Sciences
  (ICCAIS)}.\hskip 1em plus 0.5em minus 0.4em\relax IEEE, 2022, pp. 385--390.

\bibitem{kaess2012isam2}
M.~Kaess, H.~Johannsson, R.~Roberts, V.~Ila, J.~J. Leonard, and F.~Dellaert,
  ``isam2: Incremental smoothing and mapping using the bayes tree,'' \emph{The
  International Journal of Robotics Research}, vol.~31, no.~2, pp. 216--235,
  2012.

\bibitem{grisetti2005improving}
G.~Grisetti, C.~Stachniss, and W.~Burgard, ``Improving grid-based slam with
  rao-blackwellized particle filters by adaptive proposals and selective
  resampling,'' in \emph{Proceedings of the 2005 IEEE international conference
  on robotics and automation}.\hskip 1em plus 0.5em minus 0.4em\relax IEEE,
  2005, pp. 2432--2437.

\bibitem{konolige2010efficient}
K.~Konolige, G.~Grisetti, R.~K{\"u}mmerle, W.~Burgard, B.~Limketkai, and
  R.~Vincent, ``Efficient sparse pose adjustment for 2d mapping,'' in
  \emph{2010 IEEE/RSJ International Conference on Intelligent Robots and
  Systems}.\hskip 1em plus 0.5em minus 0.4em\relax IEEE, 2010, pp. 22--29.

\end{thebibliography}
\end{document}